\documentclass[letterpaper, 10 pt, conference]{ieeeconf}  
\usepackage{graphicx}
\usepackage{xcolor}
\usepackage{booktabs}
\usepackage{pifont}
\usepackage{adjustbox}
\usepackage{url}
\usepackage{hyperref}

\NewDocumentCommand{\rot}{O{45} O{1em} m}{\makebox[#2][l]{\rotatebox{#1}{#3}}}

\IEEEoverridecommandlockouts                              

\title{\Large\bf
{\em TRAVERSE}: {\em T}raffic-{\em R}esponsive {\em A}utonomous {\em V}ehicle {\em E}xperience \& {\em R}are-event {\em S}imulation for {\em E}nhanced safety
}

\author{Sandeep Thalapanane, Sandip Sharan Senthil Kumar, Guru Nandhan Appiya Dilipkumar Peethambari, \\Sourang SriHari, Laura Zheng, Julio Poveda, and Ming C. Lin \\
    \thanks{The authors are with 
 A. James Clark School of Engineering and 
 Department of Computer Science, 
         University of Maryland at College Park, MD, U.S.A.
        E-mail: \{sandeept, sandip26, guruadp, sourang, lyzheng, jpoveda, lin\}@umd.edu} %
       \href{https://gamma.umd.edu/traverse}{\texttt{gamma.umd.edu/traverse}}
       }

\begin{document}

\maketitle
\thispagestyle{empty}
\pagestyle{empty}

\begin{abstract}

Data for training learning-enabled self-driving cars in the physical world are typically collected in a safe, normal environment. Such data distribution often engenders a strong bias towards safe driving, making self-driving cars unprepared when encountering adversarial scenarios like unexpected accidents. Due to a dearth of such adverse data that is unrealistic for drivers to collect, autonomous vehicles can perform poorly when experiencing such rare events.  This work addresses much-needed research by having participants drive a VR vehicle simulator going through simulated traffic with various types of accidental scenarios. It aims to understand human responses and behaviors in simulated accidents, contributing to our understanding of driving dynamics and safety. The simulation framework adopts a robust {\em traffic simulation} and is rendered using the Unity Game Engine.  Furthermore, the simulation framework is built with portable, light-weight immersive driving simulator hardware, lowering the resource barrier for studies in autonomous driving research. 
\vspace{0.2cm}


\end{abstract}


\section{Introduction}

Intelligent systems are breaking new barriers in the automotive industry with applications ranging from robo-taxis~\cite{vosooghi2019robo} to autonomous shuttle service systems~\cite{yan2023speculative}, \cite{jones2023beyond}.  Waymo one~\cite{SCANLON2021106454}, for instance, is one such commendable innovation that has deployed self-driving taxis, capable of navigating convoluted traffic and delivering appreciable levels of safety when dealing with other vehicles on the road, compared to human drivers~\cite{deemantha2019autonomous}.  Despite promising levels of safety and precautions on the road, the autonomous industry has yet to systematically explore how such intelligent systems behave when encountered with an adversarial event, like an accident. Apart from driving behavior, external factors like jaywalking, weather conditions, animal crossing, and fortuitous behavior~\cite{abeysirigoonawardena2019generating} of neighboring vehicles can greatly engender such unforeseen events.

\begin{figure}[t]
\centering
\includegraphics[width=\columnwidth, height=6cm]{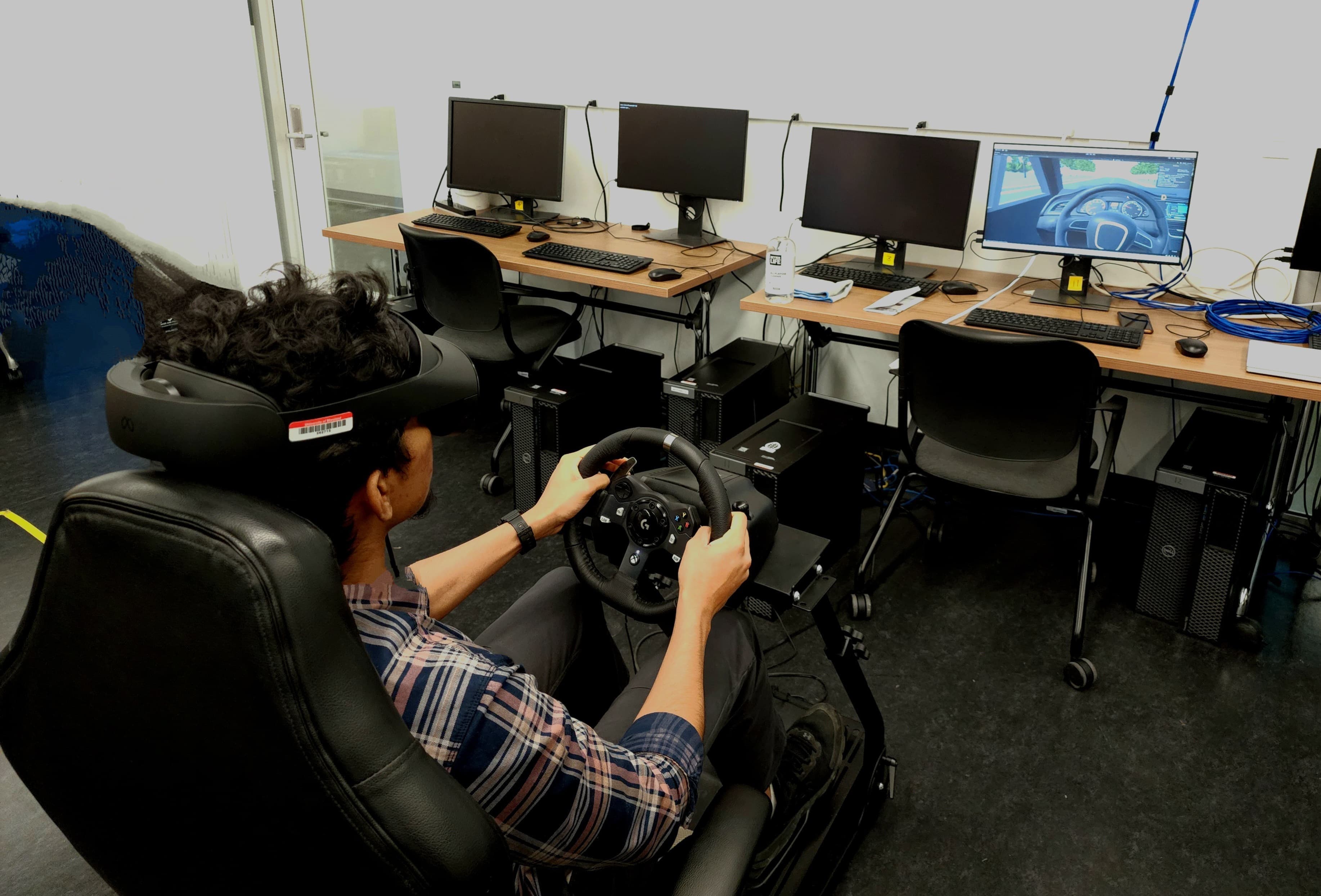}
\caption{\small{Users engage in VR driving simulation using Meta Quest Pro VR and Logitech steering wheel, with the user's view displayed on the monitor and the HMD.}}
\label{fig:User_Driving}
\end{figure}

Current methods in autonomous driving involve multiple sensors, such as stereo cameras, as well as LiDAR and Radar for mapping the surrounding environment and analyzing object orientation in three-dimensional space. 
While existing datasets are high-quality, large, and well-benchmarked within the community, they are overwhelmingly representative of normal driving scenarios. 
To address the lack of dangerous scenarios within driving datasets, current methods mostly use simulation to synthetically create traffic scenarios. 
However, such approaches are limited and barely data-driven, with risks introduced either adversarially or trivially based on assumption~\cite{slob2008state}.
Collecting data for risky scenarios is also impractical and unsafe (for human drivers) in the real world, and even less so at a large scale. 
Previous methods are limited to being data-less, simply due to the difficulty of obtaining driving data in risky scenarios safely in the real world. 

\begin{figure*}
    \centering
    \includegraphics[width=0.9\textwidth]{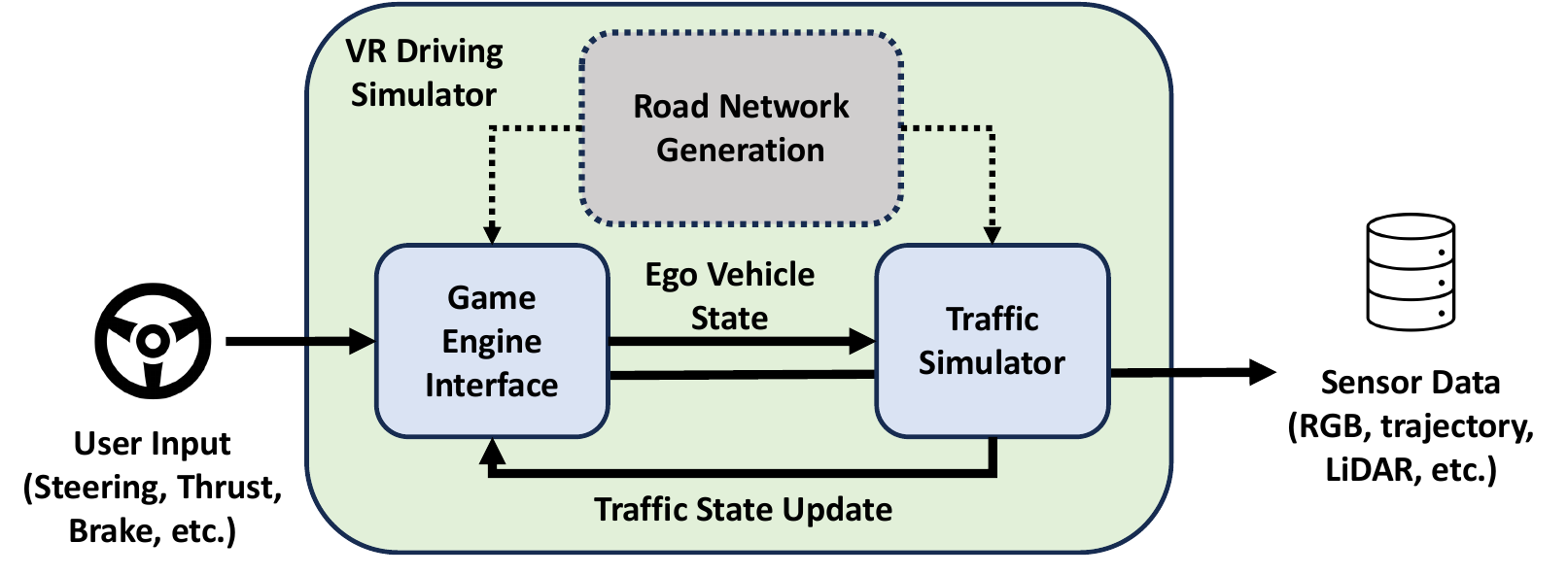}
    \caption{\textbf{Simulator System Diagram. } A schematic representation depicting the integrated flow of the simulator, encompassing the three software components: Road Network generation, Traffic simulation, and 3D simulation, aimed at enhancing the user study experience. This framework provides a comprehensive environment for simulating and studying driving behaviors and scenarios}
    \label{fig:simulator_diagram}
\end{figure*}

The other approach is to collect data in the synthetic or virtual world. 
With sufficient realism and immersion of humans in a virtual traffic environment, collecting driving data on controlled scenarios becomes possible.  The issue, however, is the lack of platform options in the simulation of realistic and immersive risky driving scenarios.  Existing simulators offer either environment realism or immersion, but not both. 

In this paper, we introduce a platform to address the lack of immersive and realistic pre-crash scenario replication. Driving simulators, like the NVIDIA Drive Sim~\cite{NVIDIA} are capable of producing higher levels of realism using VR and multi-GPU closed-loop simulations emulating actual driving experiences. 
Popular driving simulators rely on the flat screen as the optical interface~\cite{10.1145/3027063.3053202}, compromising crucial factors like immersion and field of view, due to high simulation setup cost. 
Lack of immersion may contribute to nontrivial changes in driving behavior, especially in the presence of risks. 
This can potentially truncate visual fidelity, and hence produce inaccurate data. The study proposes using Meta VR as the visual interface, compounded with Logitech Steering Wheel to ensure complete immersion and an enjoyable driving experience for users.

\begin{figure*}[t]
\centering
\includegraphics[width=0.9\textwidth, height=8cm]{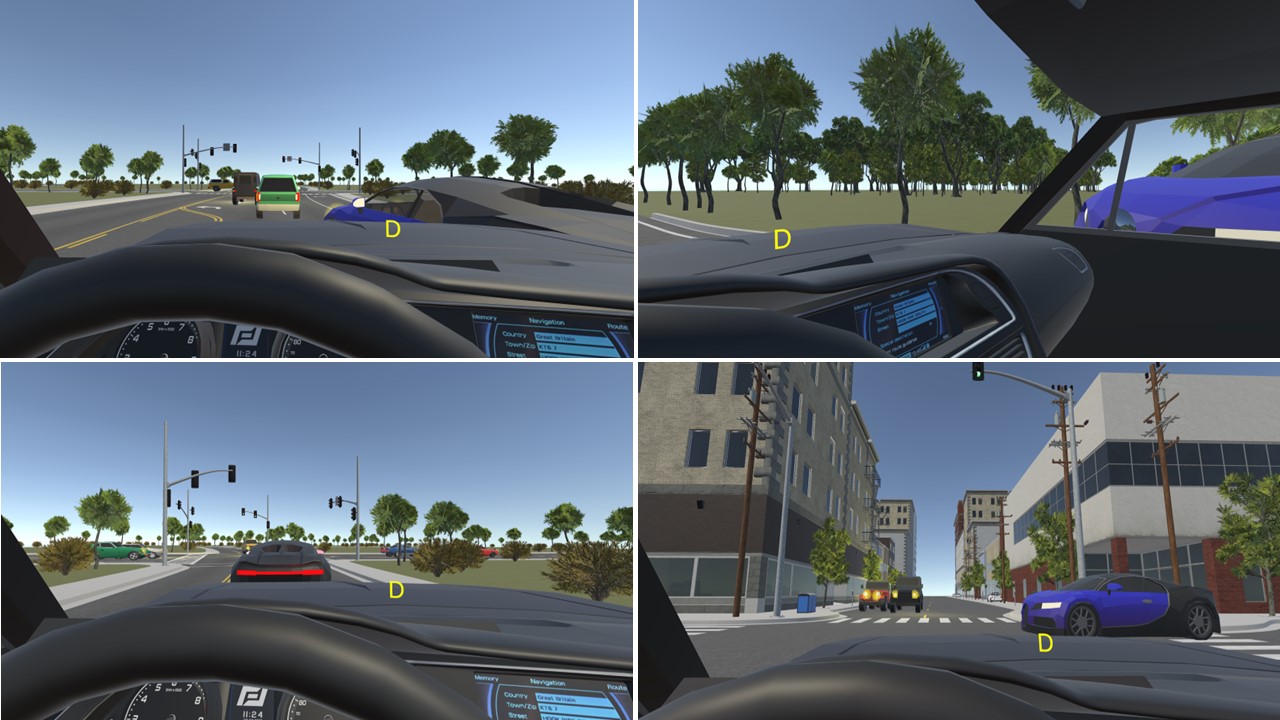}
\centering
\includegraphics[width=0.9\textwidth, height=8cm]{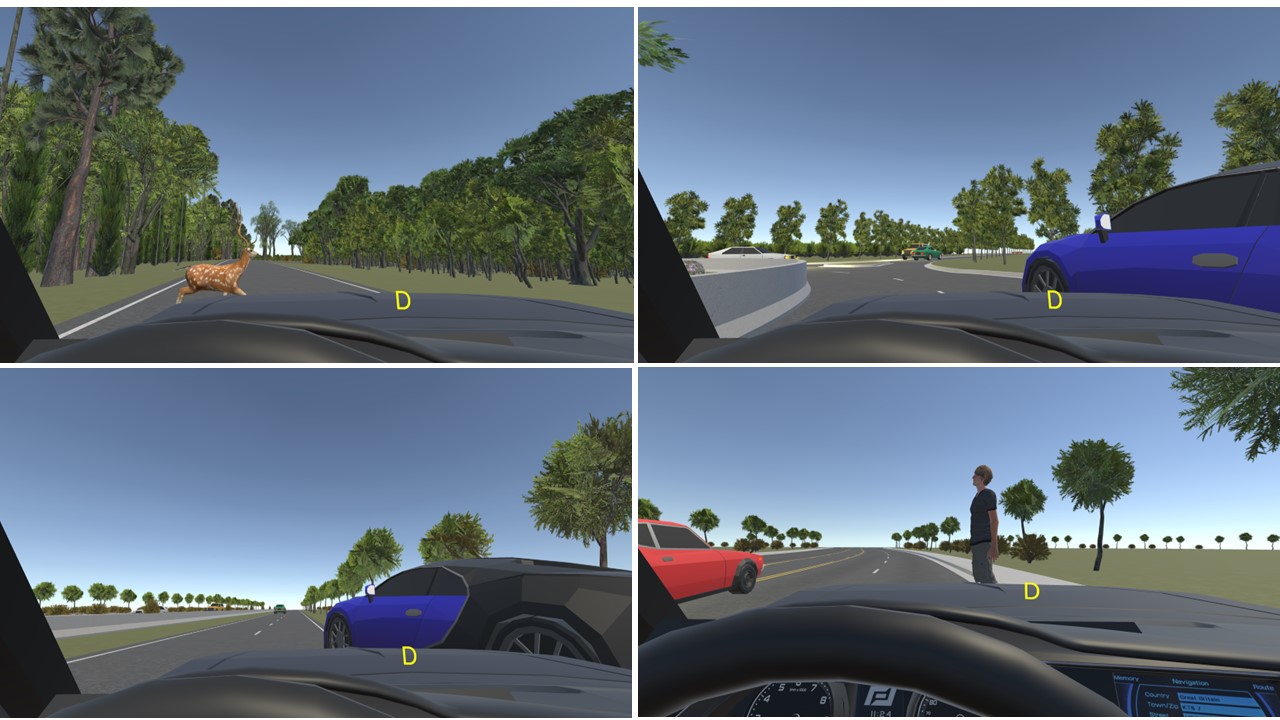}
\caption{\small{\textbf{NHTSA pre-crash scenarios implemented in our virtual reality driving simulation.} Scenarios include Sudden Lane Change Interaction, T-Bone Crash, Sudden Vehicle Stop in Front, Vehicle Running Red Lights, Sudden Deer Crossing, Crash at Roundabout, Crash in Ramp Merger, and Jaywalking Pedestrian Crash (from left to right, top to bottom). These scenarios were designed to simulate various challenging driving situations mentioned in NHTSA pre-crash typology~\cite{najm2007definition} to enhance the realism and effectiveness of the simulation}}
\vspace*{-1.5em}
\label{fig:Scenarios}
\end{figure*}

In the present study, we demonstrate a fully immersive VR-based driving simulator to generate adversarial data from a comprehensive user study using actual human drivers. We introduce 9 accident scenarios inspired by NHTSA pre-crash typology \cite{najm2007definition}. 



We present three main contributions in this paper:

(1) The proposed simulator implements {\em realistic} and {\em interactive} traffic simulation, allowing users to interfere with or influence the flow of traffic within driving scenarios. This can be useful for determining long-horizon causes and effects of fine-grain driver decisions, with use cases in traffic engineering, in addition to autonomous driving. 

(2) Our simulation platform is more portable and lightweight compared to other immersive simulators, allowing for higher frame rates, enhanced user immersion, and reducing the cost barrier for end users. 

(3) This simulation platform systematically implements scenarios from the National Highway Transportation Safety Association (NHTSA) defined pre-crash scenarios \cite{najm2007definition} and allows for easy customization out-of-the-box, enabling others to conduct flexible extensions of this research. 

We conduct user studies to validate and quantify the effects of our contribution in terms of user experience and realism, both of which are directly associated with data quality.

\section{Related Works}
In this section, we discuss the various types of existing driving simulators classified based on the level of immersion, i.e. fidelity, they provide. 

\subsection{High Fidelity Driving Simulators}

These driving simulators provide the highest level of realism to the users so that the users would react the same way as they would drive their own vehicle. The most prominent simulator is the NADS1 Driving Simulator \cite{chen2001nads}, \cite{haug1990feasibility}, \cite{10179024}. It is one of the world’s highest fidelity simulators with 13 degrees of freedom, 360-degree horizontal view, and 40-degree vertical view through 16 projectors to render the visuals. HDS~\cite{molino2005validate} by Federal Highway Administration has a 6 degrees of freedom motion-base to emulate the real-world settings, with a 200-degree display with 3 high definition projectors for visual rendering. It also consists of three LCD displays for simulating the side view and rear view mirrors. The WTI~\cite{kelly2007high} Simulator is second to the NADS simulator in terms of fidelity. The WTI Simulator consists of a 6 degrees of freedom motion base with a 240-degree FoV and a 60-degree rear view using a flat screen. The Daimler-Benz Driving Simulator~\cite{zeeb2010daimler} consists of a 360-degree screen, fast electric drive, and a twelve-meter-long rail for transverse and longitudinal movements with a maximum vehicle speed of 10m/s. All the simulators mentioned above can hold a vehicle cab for controlling the vehicle in the simulator and also have ambient environment audio cues for better immersion. Although these simulators provide a better level of realism, the issue persists in the cost of building the simulator and their low portability.

\subsection{Low Fidelity Driving Simulators}

To address the major issue of the high-fidelity simulators on accessibility and portability, the low-fidelity simulators are introduced. Mostly these low-fidelity simulators are flat screen-based simulators, making them more portable and reducing the cost of the simulator as a whole while compromising the level of realism.
The University of IOWA created a simpler version of the simulator called miniSim~\cite{miniSim}. The miniSim simulator consists of three displays to get a field of view of 180 degrees and the major features of the simulator are its support for eye tracking, video recording, and an infotainment system. The user controls the vehicle using a steering wheel setup with a haptic chair to improve immersion levels by providing ample feedback.  The AirSim~\cite{shah2018airsim},~\cite{khambhayata2023comparative} by Microsoft for drone simulation can also be used for controlling cars, which supports an object segmentation view built using Unreal Engine as a Physics Engine. 

\begin{figure*}[t]
\centering
\includegraphics[width=0.9\textwidth, height=4cm]{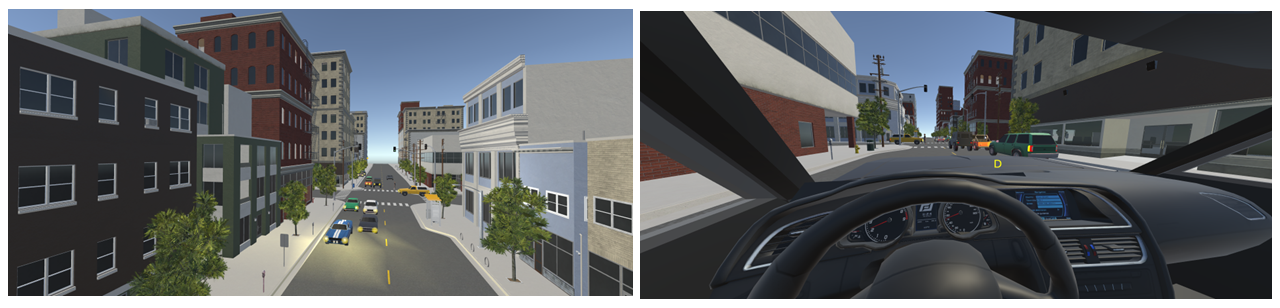}
\vspace*{-1em}
\caption{\textbf{\small{First-person and Third-person perspectives.}}{\small 
 \, (from left to right) of the simulator featuring rear-view and side-view mirrors, a speedometer, and driving modes (D/R) in a city environment with simulated traffic vehicles, and pedestrians. This view provides a realistic representation of the driving experience, enhancing the immersion for users}}
\vspace*{-1em}
\label{fig:Simulator_image}
\end{figure*}

Another simulator Deep Drive~\cite{wijaya2022deepdrive} is specifically designed for self-driving AI and is compatible with any PC. This simulator is also built on top of Unreal Engine for realistic simulation. The prominent simulator in the field of autonomous driving is CARLA~\cite{dosovitskiy2017carla}, built using Unreal Engine as a backbone.  It emulates realistic environments with elaborate urban layouts, dynamic traffic, and weather conditions, aiding testing and validation for collecting data pertaining to the perception and planning of the autonomous vehicle. CARLA supports traffic simulation through a behavior tree-based action mode called Scenario Runner~\cite{gomez2021train}, which is used to design and execute 'scenarios' with dynamic triggers and environmental effects for vehicles to navigate. Another driving simulator LGSVL~\cite{rong2020lgsvl} is built using the Unity game engine as a physics model. The simulator extends its compatibility to ROS~\cite{quigley2009ros}, using the Unity game engine plugin. The highlights of this simulator are its fish-eye camera, depth camera, LiDAR, and RADAR. Though the low-fidelity simulators provide a huge impact on reducing the cost of developing a driving simulator, they compromise the level of realism, frame rates, and immersion level of the users.

\vspace{-0.2cm}
\subsection{VR Driving Simulators}

To solve the drawbacks of both types of simulators, VR driving simulators were introduced to provide a better sense of immersion than the low-fidelity simulators and better portability and lower cost than high-fidelity simulators.

Most recently, Silvera and Biswas released an open-source, CARLA-based VR simulator called DrEyeVR~\cite{silvera2022dreyevr}.
Integrating CARLA~\cite{dosovitskiy2017carla} with VR, this simulator is capable of performing functionalities, including eye tracking, customizable traffic routes, and even ROS compatibility. The most pivotal feature is the head pose and eye gaze estimation, which plays a vital role in behavioral analysis. Other than DReyeVR there are few other VR driving simulators like Coupled Simulator~\cite{bazilinskyy2020coupled}, NVIDIA Drive Sim~\cite{NVIDIA} provides the same 360-degree FoV with better ambient environmental sounds similar to DReyeVR but lacks other features, such as eye-gaze tracking, traffic simulation, traffic signs, etc. 
Two significant limitations of the prevailing DReyeVR simulator are its limited customizability and low frames per second (FPS), resulting in challenging vehicle control and occasional discomfort for users post-VR study. In response to these shortcomings in existing driving simulators, we have introduced an innovative driving simulator. This simulator seamlessly integrates SUMO~\cite{lopez2018microscopic} and Unity~\cite{gonzalez2018unity} to provide a more authentic driving experience, coupled with immersive Virtual Reality (VR) integration. Notably, our solution addresses these challenges by offering enhanced customization options and a substantially higher frame rate, ensuring improved user control and minimizing the risk of discomfort during and after VR studies.


\section{Driving Simulator}


\subsection{Simulator Requirements}
\begin{itemize}
    \item \textbf{Interactive Traffic Simulation}: The simulator should include a realistic and interactive traffic simulation, giving users the ability to affect traffic flow in driving scenarios. This feature helps analyze how small driver decisions can impact traffic patterns over time.
    
    \item \textbf{Customizable Scenario Creation}: Road layouts need to be implemented to facilitate the creation of desired roadway scenes, allowing for road models designed to create various traffic scenarios (e.g. pre-crash).

    \item \textbf{Server Connectivity}: The simulator must allow the integration of additional controller setups to govern the behavior of accident-inducing vehicles.  Thus, it requires to establish a client-server architecture to enable data transmission from a secondary computer to the host.

    \item \textbf{Weather Conditions, Day/Night Simulation}: The simulator must have capabilities to simulate weather conditions (e.g. rain, fog, etc) and times of day (evening, night, dawn settings). These features directly affect the quality of perception. 

    \item \textbf{Functional Vehicle Features}: These include functional brake lights, turn signals, and headlights that adjust based on the environment -- critical for visual fidelity and immersive driving experience.

    \item \textbf{ROS Compatibility}: 
    This compatibility enables users to leverage the simulator within the broader ROS ecosystem (ROS and ROS2), facilitating the development, testing, and deployment of robotic/autonomous systems.

    \item \textbf{Eye Gaze, Head Pose, \& Hand Tracking}: 
    is required to monitor the user's eye gaze, head pose, and hand position in the simulated vehicle.

    \item \textbf{Recording Gameplay}: The simulator must have the capability of recording human subject’s simulations which can be employed to analyze the effectiveness of the study in terms of realism and immersion. 

\end{itemize}


\subsection{Design Implementation}

\subsubsection{Hardware Components}
Our system uses the following devices in our prototype implementation.
\begin{itemize}

\item \textbf{Meta Quest Pro VR}: Selected for its eye-gaze tracking functionality, which allows for post-simulation research. The device offers a high level of immersion for users.

\item \textbf{Logitech G29 steering wheel}: Chosen for its pedal kit and precise control, enabling users to control acceleration, braking, steering, and gear shifting, enhancing haptic feedback of the driving experience.
\item \textbf{PC}: a Windows 11 x64 system with Intel Xenon(R) Gold 5218 CPU (x2 Processor),
64GB RAM and Nvidia GeForce RTX 2080 Ti with 2TB SSD\\
\end{itemize}


\vspace{-0.4cm}
\subsubsection{Softwares}
We implemented our immersive driving simulator using the customized, enhanced, and integrated implementation of the software listed below.
\begin{itemize}
\item \textbf{RoadRunner}: Used for creating detailed and realistic 3D scenes with complex road networks and landscapes. RoadRunner~\cite{acharyalateral},~\cite{he2018roadrunner} offers a user-friendly interface for designing road layouts, terrain features, and structures.

\item \textbf{SUMO (Simulation of Urban MObility)}: Enhances driving scenario realism by modeling vehicle movement, traffic flow, and intersections. SUMO provides a microscopic simulation of continuous, multimodal vehicle traffic in large urban road networks~\cite{lopez2018microscopic},~\cite{kaths2016integrated}.

\item \textbf{Unity Game Engine}: Integrated with the road maps created in RoadRunner with the dynamic 2D traffic scenarios simulated by the traffic simulator as shown in Figure \ref{fig:simulator_diagram}. Unity provides a powerful platform for creating immersive simulations~\cite{gonzalez2018unity},~\cite{szalai2020mixed},~\cite{biurrun2017microscopic},~\cite{yang2016unity}.
\end{itemize}

\begin{figure*}[!t]
\centering
\includegraphics[width=0.8\textwidth]{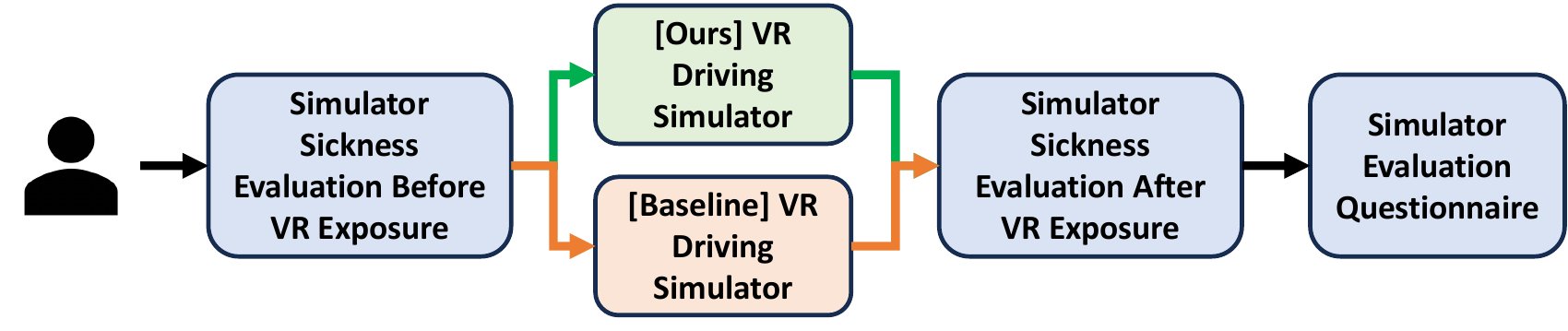}
\vspace*{-1em}
\caption{\textbf{User study design for systematic evaluation of VR driving simulation quality. } The user study is evaluated in two stages (green and orange arrows). Each evaluated participant will complete simulator sickness evaluations before and after each stage, in addition to a simulator evaluation questionnaire (black arrows). In both stages, the participant drives through NHTSA pre-crash scenarios of each simulator and evaluates the quality in several dimensions, further detailed in Section~\ref{sec:user_study_description}. The order at which the user experiences each simulator is randomized during the study to account for order bias. Additionally, users complete each stage on two separate days, in order to minimize sickness effects carrying over to the second stage.}
\vspace*{-1em}
\label{fig:framework}
\end{figure*}

\begin{figure*}[t]
\centering
\includegraphics[width=0.9\textwidth]{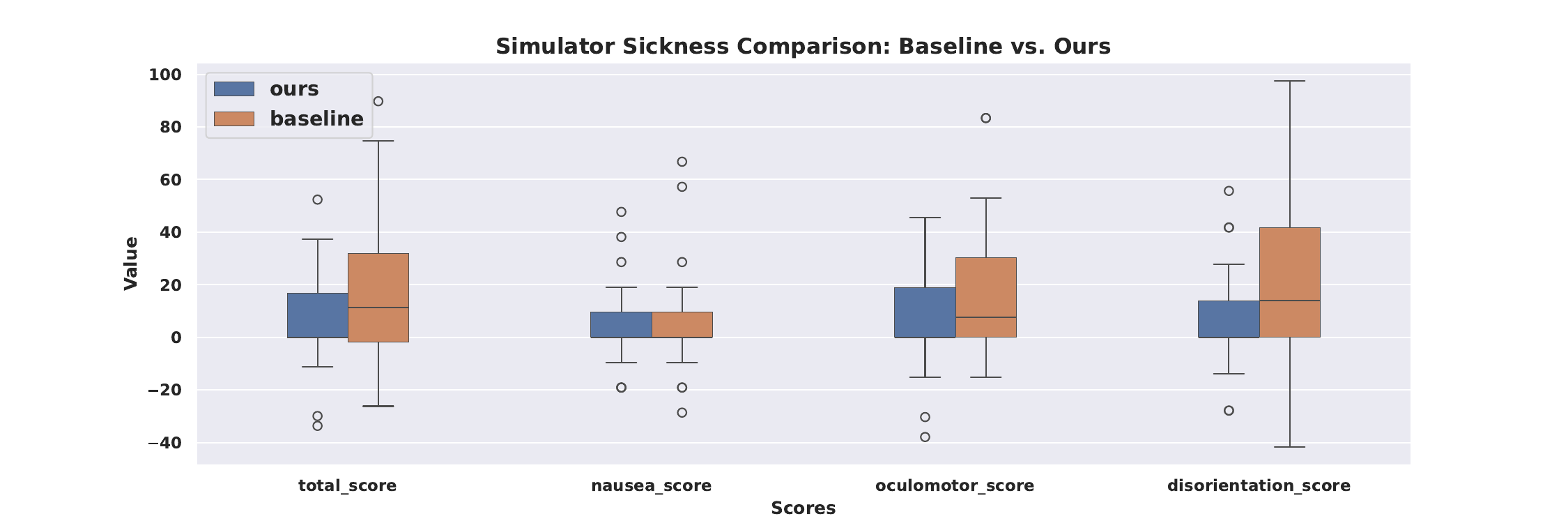}
\vspace*{-1em}
\caption{\small{{\bf Sickness levels rated by participants after driving each simulator}, measured on a scale of ten and averaged for analysis. The graph illustrates the comparative simulator sickness~\cite{somrak2019estimating} experienced by participants in our simulator versus the baseline simulator on aspects such as total score, nausea score, oculomotor score, disorientation score (from left to right).  Overall, the participants experienced {\bf less} motion sickness when driving our simulator.}}
\vspace*{-1em}
\label{fig:sickness_scale}
\end{figure*}

\begin{figure}[t]
\centering
\includegraphics[width=\columnwidth, height=3cm]{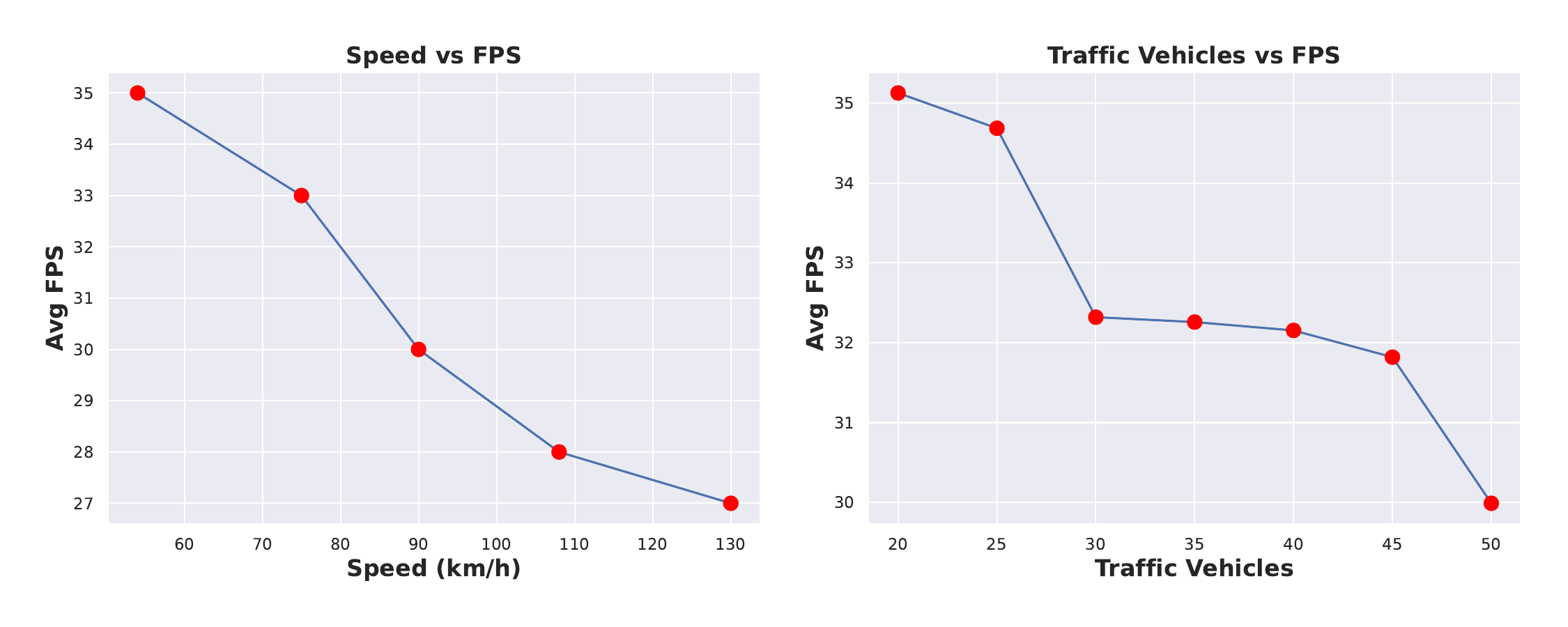}
\vspace*{-2em}
\caption{\small{{\bf Changes in frame rates.} The graphs show the relationship between FPS and the number of vehicles (left) and the relationship between FPS and the maximum speed of the ego car (right).}}
\vspace*{-1em}
\label{fig:Simulator_Analysis}
\end{figure}

\section{Data Collection}

The primary aim is to devise a range of diverse and challenging accident scenarios for simulation, drawing inspiration from the NHTSA pre-crash typology. These scenarios seek to elicit varying responses from human drivers, revealing their adaptability and decision-making prowess in minimizing potential impacts. Eight scenarios are shown in Figure~\ref{fig:Scenarios}, with one additional scenario designated as a practice scenario to familiarize users with vehicle controls. 
 Participants would be presented with a randomized sequence comprising eight scenarios per driver, mitigating any inherent order bias. Each scenario is designed to last approximately 1 to 3 minutes, providing an adequate time-frame for a comprehensive assessment of participants' reaction times across different accident scenarios. 
 Data collection entails the integration of hardware components to enhance user engagement and gather crucial insights into system responses. This process incorporates the utilization of a Logitech steering wheel for controlling the ego vehicle and MetaVR technology to facilitate an immersive virtual reality environment.

\begin{table}[!t]
\centering
\caption{\small{{\bf Comparison of varying criteria between two VR driving simulators.} The table presents t-statistic, p-value, and indicates there is a statistical significance between our vs. baseline simulators for each criterion.}}
\begin{tabular}{l|r|r|c}
\toprule
Criterion & T-statistic & p-value & Reject $H_0$ \\ 
\midrule
Sense of Being in VR          & 3.895                & 2.49e-4        & Yes                                       \\ 
Ease of adjustment            & 5.748                & 3.22e-7         & Yes                                       \\ 
Scenario Realism              & 3.164                & 2.45e-3        & Yes                                       \\
Controls Responsiveness       & 4.476                & 3.41e-5         & Yes                                       \\ 
Audio Immersiveness           & 2.046                & 4.52e-2         & Yes                                       \\ 
Head Tracking                 & 4.801                & 1.08e-5         & Yes                                       \\
Traffic Simulation            & 0.620                & 5.37e-1         & No                                        \\
Realistic Control   & 4.238                & 7.89e-5         & Yes                                       \\
Overall Experience            & 5.595                & 5.76e-7         & Yes                                       \\  
\bottomrule
\end{tabular}
\label{tb:T-Test}
\end{table}

\subsection{Scenario-Based Assessments}

The collection of data parameters during these scenarios is comprehensive. The system records timestamps, capturing the temporal aspects of events during the simulation. The spatial location of the ego vehicle is tracked through car coordinates, while brake usage provides insights into the user's reactions to sudden stops or collision scenarios. Additionally, the measurement of the steering angle from the Logitech steering wheel indicates user responses to scenarios involving lane changes, obstacles, or complex traffic situations. The Logitech steering wheel, renowned for its precision and responsiveness, captures subtle user inputs such as steering angle, gas, and brake application, as well as vehicle acceleration for trajectory analysis. Similarly, MetaVR facilitates the collection of user eye gaze position, eye openness, and blink frequency, enabling detailed post-analysis of participant behavior. Figure \ref{fig:User_Driving} shows the experiment setup for the user study, and Figure~\ref{fig:framework} shows the study design for the systematic evaluation of VR driving simulators.


\begin{figure*}[t]
\centering
\includegraphics[width=0.95\textwidth]{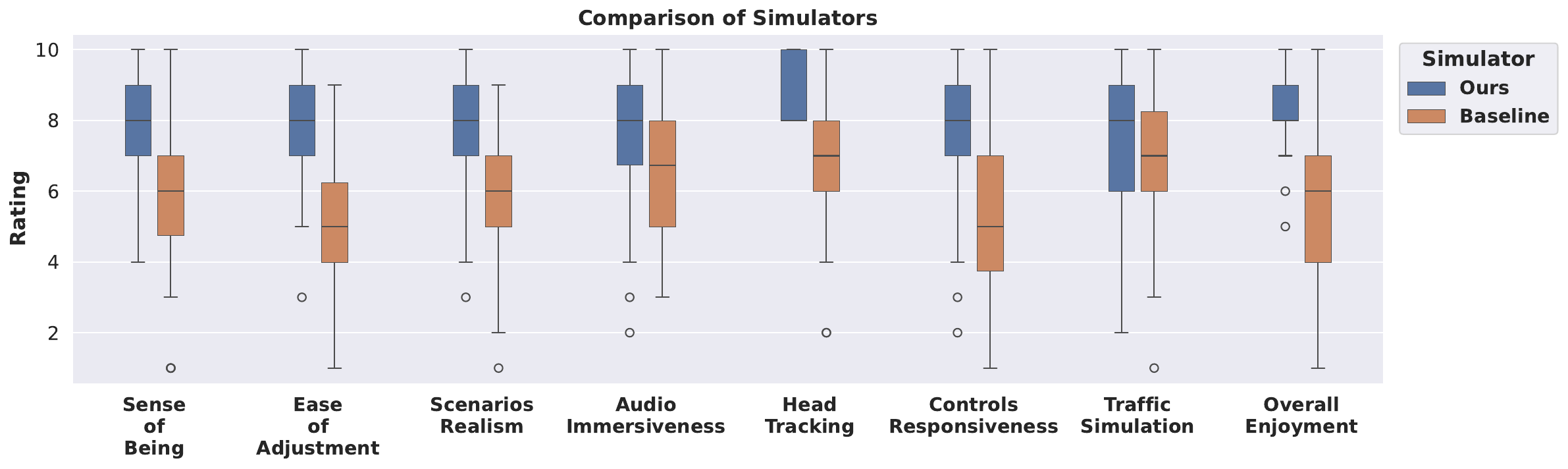}
\caption{\small{{\bf Comparison on ratings by the users on different aspects of the simulators.} The graph displays the comparison of user ratings between our simulator and the baseline simulator across various aspects, including sense of being, ease of adjustment, scenarios realism, audio immersiveness, head tracking, control responsiveness, traffic simulation and overall enjoyment (from left to right).  Our immersive driving simulator is consistently rated higher than the SOTA driving simulator across all aspects.}}
\label{fig:Individual_ratings}
\end{figure*}

\begin{figure*}[t]
\centering
\includegraphics[width=0.95\textwidth]{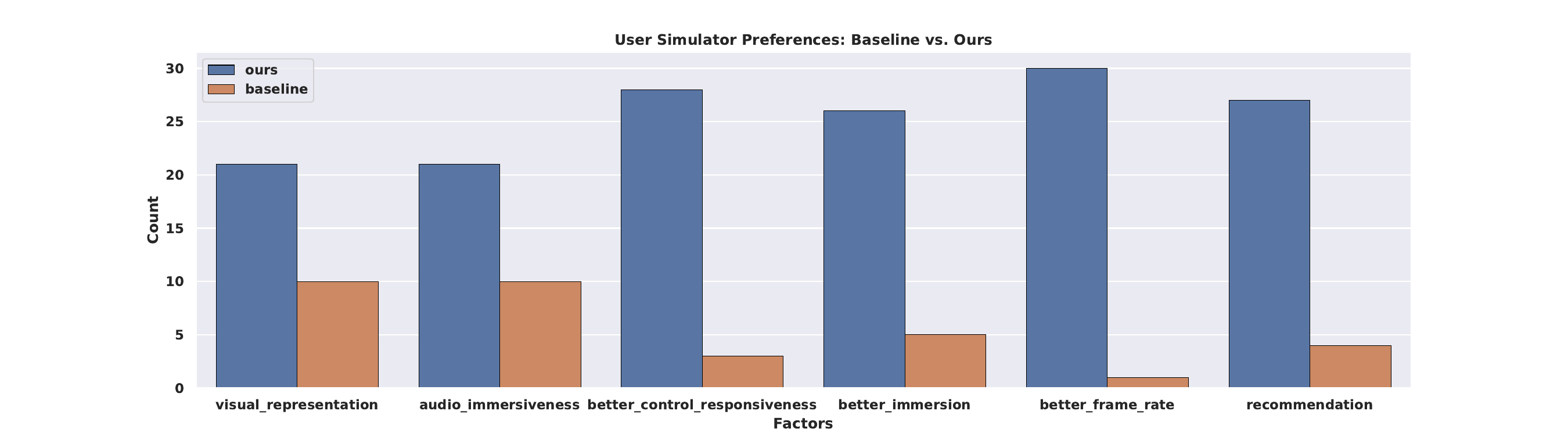}
\caption{\small{{\bf User evaluation of the simulators across multiple dimensions} (from left to right): visual representation, audio immersiveness, better control responsiveness, better immersion, better frame rate, recommended.
The chart depicts the comparative analysis of user preferences between our simulator (consistently higher across all dimensions) and the baseline SOTA driving simulator.}}
\label{fig:User_preferences}
\end{figure*}

\subsection{User Study Evaluation on Simulation Fidelity}
\label{sec:user_study_description}
Recognizing the multifaceted nature of evaluating a simulator's effectiveness, which involves both quantitative measurements and subjective user perceptions, our objective was to attain a thorough understanding by conducting a comparative user study with 31 participants. This study aimed to gather comprehensive insights by comparing our simulator with the baseline simulator (DReyeVR)~\cite{silvera2022dreyevr}. Each participant was allotted five minutes to operate in each simulator and subsequently completed a pre-simulator, VR simulator, and post-simulator questionnaires. To mitigate potential carryover effects, participants were granted a minimum one-hour interval between successive simulator sessions, ensuring any adverse effects experienced in the first session did not influence the second. To counteract the potential order bias, participants were randomly assigned one of two simulators, with some receiving our simulator as their first experience and the remainder receiving the baseline simulator first. Figure \ref{fig:framework} represents the whole process of the user study. The pre-simulator questionnaire solicited optional demographic data and self-assessments of physical well-being and susceptibility to motion sickness. Following the VR experience, the post-study questionnaire elicited similar self-assessments of well-being and sickness, alongside an evaluation of VR system immersion. Finally, users are asked to fill out a final comparison questionnaire which requests for the user's preferred simulator based on aspects such as visual representation, audio representation, better control responsiveness, better immersion, better frame rate, and recommendations to other users.



\section{Results}

\subsection{Qualitative Comparison To Existing Simulators}


The simulator has a fidelity score of 9/15 points according to the simulator fidelity scoring system designed by Meta review of Wynne et al~\cite{WYNNE2019138}. This strongly indicates that our driving simulator belongs to the medium simulator fidelity category, provided that it was built with low cost and limited hardware capabilities.  The distribution of the score involves a point for a no-motion base, five points for 360\textdegree \, the field of view in terms of visuals, and three for the Logitech steering wheel including an arcade seating mechanism, a total of nine points. The fidelity score increases with augmented levels of immersion and realism by incorporating convoluted hardware devices and setup configurations. For instance, a flat-screen driving simulator~\cite{10.1145/3027063.3053202} would have a low score of six, whereas architectures like CAVE visual systems~\cite{cruz1992cave} with a driver cabin would score 11 points.

The data collection process during the VR simulation encompasses distinct driving features including hand and eye tracking, speedometer, and different mirror views of the vehicle as depicted in Figure \ref{fig:Simulator_image}. Conversely, there are features that are exclusive to this study and have not been adopted by contemporary autonomous driving studies for the purpose of complete immersion and realism. Eye openness is one such feature that determines the degree of how wide open the eyelids of the driver are while driving. This feature is crucial in determining the level of alertness as sleeping while driving is a major cause of accidents. The simulator architecture facilities incorporate additional scenarios which can be beneficial to add uncaptured accident data.  Finally, the most innovative component of the overall immersive experience involves operating two vehicles in the same scenario. Traffic planning of other vehicles can only possess a certain level of veracity and this is augmented by incorporating another user in the simulator which is revolutionary for our analysis as elements like realism and driver-to-driver interaction augment to a great extent.


\subsection{Quantitative Pilot Study Evaluations}

In total, 31 participants were administered a questionnaire aimed at evaluating the optimal simulator based on criteria encompassing Visual and audio immersiveness, control responsiveness,  immersiveness, and Frames Per Second (FPS). The sickness level scores~\cite{somrak2019estimating} in Figure \ref{fig:Simulator_Analysis} show that although our simulator exhibits similar levels of nausea, the overall scores including oculomotor and disorientation scores are lower when compared to the baseline SOTA simulator. The user experience segment of the user study, involving features like sense of being, enjoyment, realism, and ease of adjustment is calibrated as part of the study, and it can be visible in Figure~\ref{fig:sickness_scale}: {\em all these simulator-oriented experiences are much better in our simulator}. User experience is very important to the study, as it validates the dataset collected and determines the realism of the study -- which has been delivered in all aspects successfully. 

The statistical analysis was carried out to investigate the significance of the comparison aspects of the simulator. The t-test~\cite{kim2015t} was performed on the user responses of both the simulators and the obtained t-statistic and p-value were recorded. We choose a threshold value of p=0.05 to determine statistical significance. The results indicated a significant difference in most of the aspects of the simulator, except for the traffic simulation. The summarized information is presented in Table \ref{tb:T-Test}. 
We believe the inconclusive results in traffic simulation may be due to the nature of the scenarios itself, as users may be less likely to rate traffic as realistic upon experiencing a risky scenario.

It is noteworthy, as depicted in Figure \ref{fig:Individual_ratings}, that the frame rate and vehicle speed exhibit an inverse relationship. Maintaining a smooth transition of frames, alongside an optimal vehicle speed, is imperative for replicating real-world driving conditions. According to our analysis, the simulator's frames per second (FPS) decreased to 30 when the number of vehicles reached 50, a level at which users are unlikely to experience discomfort while driving. The majority of users expressed a preference for our simulator across all criteria, attesting to its fidelity to real-world driving scenarios which is shown in Figure \ref{fig:User_preferences}.  A high score of visual representation and audio immersiveness augments the veracity of the simulator, which are clear indicators of a real-world driving experience and surrounding environment.



\section{Conclusion}

By leveraging high-fidelity 3D game engine, customizing detailed traffic simulation with realistic road networks, and portable, light-weight VR display and driving hardware, we create a comprehensive, accurate evaluations of autonomous driving systems, accompanied by rare-event simulation for collecting driving personality data under varying pre-crash conditions for enhanced safety of self-driving cars. This combination provides a versatile framework for simulating diverse driving scenarios and testing algorithms and strategies for autonomous vehicles, ultimately advancing the development and deployment of autonomous driving technology for better safety. We also aim to make the simulator differentiable by integrating differentiable numerical simulations (or differentiable physics), enabling seamless learning and control, with potential applications in training self-driving cars to navigate safely in complex environments. Leveraging the data acquired from the user study, we endeavor to potentially classify the driving behaviors of surrounding vehicles for motion and trajectory prediction of nearby vehicles. This information can then be harnessed to enhance the capabilities of self-driving cars, enabling them to navigate roads with heightened caution and precision.

\bibliographystyle{ieeetr}
\bibliography{references}

\begin{thebibliography}{10}

\bibitem{vosooghi2019robo}
R.~Vosooghi, J.~Kamel, J.~Puchinger, V.~Leblond, and M.~Jankovic, ``Robo-taxi service fleet sizing: assessing the impact of user trust and willingness-to-use,'' {\em Transportation}, vol.~46, no.~6, pp.~1997--2015, 2019.

\bibitem{yan2023speculative}
M.~Yan, Z.~Lin, P.~Lu, M.~Wang, L.~Rampino, and G.~Caruso, ``Speculative exploration on future sustainable human-machine interface design in automated shuttle buses,'' {\em Sustainability}, vol.~15, no.~6, p.~5497, 2023.

\bibitem{jones2023beyond}
R.~Jones, J.~Sadowski, R.~Dowling, S.~Worrall, M.~Tomitsch, and E.~Nebot, ``Beyond the driverless car: A typology of forms and functions for autonomous mobility,'' {\em Applied Mobilities}, vol.~8, no.~1, pp.~26--46, 2023.

\bibitem{SCANLON2021106454}
J.~M. Scanlon, K.~D. Kusano, T.~Daniel, C.~Alderson, A.~Ogle, and T.~Victor, ``Waymo simulated driving behavior in reconstructed fatal crashes within an autonomous vehicle operating domain,'' {\em Accident Analysis \& Prevention}, vol.~163, p.~106454, 2021.

\bibitem{deemantha2019autonomous}
R.~Deemantha and B.~Hettige, ``Autonomous car: Current issues, challenges and solution: A review,'' in {\em Proceedings of the 15th International Research Conference, Beijing, China}, pp.~1--6, 2019.

\bibitem{abeysirigoonawardena2019generating}
Y.~Abeysirigoonawardena, F.~Shkurti, and G.~Dudek, ``Generating adversarial driving scenarios in high-fidelity simulators,'' in {\em 2019 International Conference on Robotics and Automation (ICRA)}, pp.~8271--8277, IEEE, 2019.

\bibitem{slob2008state}
J.~J. Slob, ``State-of-the-art driving simulators, a literature survey,'' {\em DCT report}, vol.~107, 2008.

\bibitem{NVIDIA}
``{NVIDIA DRIVE} sim.'' \url{https://developer.nvidia.com/drive/}, 2018.

\bibitem{10.1145/3027063.3053202}
M.~Walch, J.~Frommel, K.~Rogers, F.~Sch\"{u}ssel, P.~Hock, D.~Dobbelstein, and M.~Weber, ``Evaluating vr driving simulation from a player experience perspective,'' in {\em Proceedings of the 2017 CHI Conference Extended Abstracts on Human Factors in Computing Systems}, CHI EA '17, (New York, NY, USA), p.~2982–2989, Association for Computing Machinery, 2017.

\bibitem{najm2007definition}
W.~G. Najm and D.~L. Smith, ``Definition of a pre-crash scenario typology for vehicle safety research,'' in {\em Proceeding of the 20th international technical conference on the enhanced safety of vehicles}, 2007.

\bibitem{chen2001nads}
L.~Chen, Y.~Papelis, G.~Waston, and D.~Solis, ``Nads at the university of iowa: A tool for driving safety research,'' in {\em Proceedings of the 1st human-centered transportation simulation conference}, 2001.

\bibitem{haug1990feasibility}
E.~J. Haug, ``Feasibility study and conceptual design of a national advanced driving simulator. final report,'' tech. rep., 1990.

\bibitem{10179024}
C.~Schwarz, O.~Ahmad, T.~Brown, J.~Gaspar, G.~Wagner, and D.~V. McGehee, ``The long and winding road: 25 years of the national advanced driving simulator,'' {\em IEEE Computer Graphics and Applications}, vol.~43, no.~4, pp.~121--128, 2023.

\bibitem{molino2005validate}
J.~Molino, B.~Katz, D.~Duke, K.~Opiela, C.~Andersen, and M.~Moyer, ``Validate first; simulate later: a new approach used at the fhwa highway driving simulator,'' in {\em Proceedings of Driver Simulation Conference, North America, Orando, FL}, pp.~411--420, 2005.

\bibitem{kelly2007high}
M.~J. Kelly, S.~Lassacher, Z.~Shipstead, {\em et~al.}, ``A high fidelity driving simulator as a tool for design and evaluation of highway infrastructure upgrades,'' tech. rep., Montana. Dept. of Transportation. Research Programs, 2007.

\bibitem{zeeb2010daimler}
E.~Zeeb, ``Daimler’s new full-scale, high-dynamic driving simulator--a technical overview,'' {\em Actes INRETS}, pp.~157--165, 2010.

\bibitem{miniSim}
``{miniSim} driving safety research institute, national advanced driving simulator, university of iowa.'' \url{https://www.nads-sc.uiowa.edu/minisim/}, 2013.

\bibitem{shah2018airsim}
S.~Shah, D.~Dey, C.~Lovett, and A.~Kapoor, ``Airsim: High-fidelity visual and physical simulation for autonomous vehicles,'' in {\em Field and Service Robotics: Results of the 11th International Conference}, pp.~621--635, Springer, 2018.

\bibitem{khambhayata2023comparative}
M.~Khambhayata, ``A comparative analysis of carla and airsim simulators: Investigating implementation challenges in autonomous driving,'' {\em Available at SSRN 4477130}, 2023.

\bibitem{wijaya2022deepdrive}
N.~Wijaya, S.~H. Mulyani, and A.~C. Noviadi~Prabowo, ``Deepdrive: effective distracted driver detection using generative adversarial networks (gan) algorithm,'' {\em Iran Journal of Computer Science}, vol.~5, no.~3, pp.~221--227, 2022.

\bibitem{dosovitskiy2017carla}
A.~Dosovitskiy, G.~Ros, F.~Codevilla, A.~Lopez, and V.~Koltun, ``Carla: An open urban driving simulator,'' in {\em Conference on robot learning}, pp.~1--16, PMLR, 2017.

\bibitem{gomez2021train}
C.~G{\'o}mez-Hu{\'e}lamo, J.~Del~Egido, L.~M. Bergasa, R.~Barea, E.~L{\'o}pez-Guill{\'e}n, F.~Arango, J.~Araluce, and J.~L{\'o}pez, ``Train here, drive there: Simulating real-world use cases with fully-autonomous driving architecture in carla simulator,'' in {\em Advances in Physical Agents II: Proceedings of the 21st International Workshop of Physical Agents (WAF 2020), November 19-20, 2020, Alcal{\'a} de Henares, Madrid, Spain}, pp.~44--59, Springer, 2021.

\bibitem{rong2020lgsvl}
G.~Rong, B.~H. Shin, H.~Tabatabaee, Q.~Lu, S.~Lemke, M.~Mo{\v{z}}eiko, E.~Boise, G.~Uhm, M.~Gerow, S.~Mehta, {\em et~al.}, ``Lgsvl simulator: A high fidelity simulator for autonomous driving,'' in {\em 2020 IEEE 23rd International conference on intelligent transportation systems (ITSC)}, pp.~1--6, IEEE, 2020.

\bibitem{quigley2009ros}
M.~Quigley, K.~Conley, B.~Gerkey, J.~Faust, T.~Foote, J.~Leibs, R.~Wheeler, A.~Y. Ng, {\em et~al.}, ``Ros: an open-source robot operating system,'' in {\em ICRA workshop on open source software}, vol.~3, p.~5, Kobe, Japan, 2009.

\bibitem{silvera2022dreyevr}
G.~Silvera, A.~Biswas, and H.~Admoni, ``Dreyevr: Democratizing virtual reality driving simulation for behavioural \& interaction research,'' 2022.

\bibitem{bazilinskyy2020coupled}
P.~Bazilinskyy, L.~Kooijman, D.~Dodou, and J.~C. de~Winter, ``Coupled simulator for research on the interaction between pedestrians and (automated) vehicles,'' in {\em Driving Simulation Conference Europe. Antibes, France}, 2020.

\bibitem{lopez2018microscopic}
P.~A. Lopez, M.~Behrisch, L.~Bieker-Walz, J.~Erdmann, Y.-P. Fl{\"o}tter{\"o}d, R.~Hilbrich, L.~L{\"u}cken, J.~Rummel, P.~Wagner, and E.~Wie{\ss}ner, ``Microscopic traffic simulation using sumo,'' in {\em 2018 21st international conference on intelligent transportation systems (ITSC)}, pp.~2575--2582, IEEE, 2018.

\bibitem{gonzalez2018unity}
D.~Gonz{\'a}lez~Ortega, F.~J. D{\'\i}az~Pernas, M.~Mart{\'\i}nez~Zarzuela, M.~Ant{\'o}n~Rodr{\'\i}guez, {\em et~al.}, ``Unity-based simulation scenarios to study driving performance,'' 2018.

\bibitem{acharyalateral}
S.~Acharya, S.~Mane, C.~Kanade, R.~Hatkar, and A.~Marathe, ``Lateral and longitudinal control of an autonomous vehicle,''

\bibitem{he2018roadrunner}
S.~He, F.~Bastani, S.~Abbar, M.~Alizadeh, H.~Balakrishnan, S.~Chawla, and S.~Madden, ``Roadrunner: improving the precision of road network inference from gps trajectories,'' in {\em Proceedings of the 26th ACM SIGSPATIAL international conference on advances in geographic information systems}, pp.~3--12, 2018.

\bibitem{kaths2016integrated}
J.~Kaths and S.~Krause, ``Integrated simulation of microscopic traffic flow and vehicle dynamics,'' in {\em IPG Apply \& Innovate 2016}, 2016.

\bibitem{szalai2020mixed}
M.~Szalai, B.~Varga, T.~Tettamanti, and V.~Tihanyi, ``Mixed reality test environment for autonomous cars using unity 3d and sumo,'' in {\em 2020 IEEE 18th World Symposium on Applied Machine Intelligence and Informatics (SAMI)}, pp.~73--78, IEEE, 2020.

\bibitem{biurrun2017microscopic}
C.~Biurrun-Quel, L.~Serrano-Arriezu, and C.~Olaverri-Monreal, ``Microscopic driver-centric simulator: Linking unity3d and sumo,'' in {\em Recent Advances in Information Systems and Technologies: Volume 1 5}, pp.~851--860, Springer, 2017.

\bibitem{yang2016unity}
C.-W. Yang, T.-H. Lee, C.-L. Huang, and K.-S. Hsu, ``Unity 3d production and environmental perception vehicle simulation platform,'' in {\em 2016 International Conference on Advanced Materials for Science and Engineering (ICAMSE)}, pp.~452--455, IEEE, 2016.

\bibitem{somrak2019estimating}
A.~Somrak, I.~Humar, M.~S. Hossain, M.~F. Alhamid, M.~A. Hossain, and J.~Guna, ``Estimating vr sickness and user experience using different hmd technologies: An evaluation study,'' {\em Future Generation Computer Systems}, vol.~94, pp.~302--316, 2019.

\bibitem{WYNNE2019138}
R.~A. Wynne, V.~Beanland, and P.~M. Salmon, ``Systematic review of driving simulator validation studies,'' {\em Safety Science}, vol.~117, pp.~138--151, 2019.

\bibitem{cruz1992cave}
C.~Cruz-Neira, D.~J. Sandin, T.~A. DeFanti, R.~V. Kenyon, and J.~C. Hart, ``The cave: Audio visual experience automatic virtual environment.,'' {\em Communications of the ACM}, vol.~35, no.~6, pp.~64--73, 1992.

\bibitem{kim2015t}
T.~K. Kim, ``T test as a parametric statistic,'' {\em Korean journal of anesthesiology}, vol.~68, no.~6, p.~540, 2015.

\end{thebibliography}

\end{document}